\documentclass[10pt,twocolumn,letterpaper]{article}

\usepackage{cvpr}              

\usepackage{graphicx}
\usepackage{amsmath}
\usepackage{amssymb}
\usepackage{booktabs}
\usepackage{color}
\usepackage{colortbl}
\usepackage[dvipsnames]{xcolor}
\usepackage{float}
\usepackage{multirow}

\usepackage[pagebackref,breaklinks,colorlinks]{hyperref}

\usepackage[capitalize]{cleveref}
\crefname{section}{Sec.}{Secs.}
\Crefname{section}{Section}{Sections}
\Crefname{table}{Table}{Tables}
\crefname{table}{Tab.}{Tabs.}

\def\cvprPaperID{*****} 
\def\confName{CVPR}
\def\confYear{2022}
\def\cvprPaperID{*****} 
\def\confName{CVPR}
\def\confYear{2022}

\begin{document}

\newcommand{\vlp}{Egocentric VLP}
\newcommand{\dataset}{EgoClip}
\newcommand{\model}{EgoNCE\xspace}
\newcommand{\eval}{EgoMCQ}
\newcommand{\epic}{EPIC-KITCHENS-100}
\newcommand{\charades}{Charades-Ego}
\newcommand{\lta}{video-text localization}
\newcommand{\mir}{Multi-Instance Retrieval}
\newcommand{\nlq}{Natural Language Query}
\newcommand{\mq}{Moment Query}
\newcommand{\pnr}{PNR Localization}
\newcommand{\OSCC}{Object State Change Classification}
\newcommand{\web}{WebVid-2M}
\newcommand{\ccweb}{CC3M+WebVid-2M}
\newcommand{\howto}{HowTo100M}

\title{Egocentric Video-Language Pretraining~@~Ego4D~Challenge~2022}  

\author{Kevin Qinghong Lin$^1$\quad Alex Jinpeng Wang$^1$\quad Mattia Soldan$^3$\quad Michael Wray$^2$\\
Rui Yan$^1$\quad Eric Zhongcong Xu$^1$\quad Difei Gao$^1$\quad Rongcheng Tu$^4$\\
Wenzhe Zhao$^4$\quad Weijie Kong$^4$\quad Chengfei Cai$^4$\quad Hongfa Wang$^4$\\
Dima Damen$^2$\quad Bernard Ghanem$^3$\quad Wei Liu$^4$\quad Mike Zheng Shou$^1$\thanks{Corresponding Author.}\\
\\
$^1$Show Lab, National University of Singapore\quad $^2$University of Bristol\\
$^3$King Abdullah University of Science and Technology\quad $^4$Tencent Data Platform\\
\tt\footnotesize \{kevin.qh.lin, yanrui6019, turongcheng, mike.zheng.shou\}@gmail.com, \{jinpengwang, zhongcongxu\}@u.nus.edu\\
\tt\footnotesize \{michael.wray, dima.damen\}@bristol.ac.uk, \{mattia.soldan, bernard.ghanem\}@kaust.edu.sa, difei.gao@vipl.ict.ac.cn\\
\tt\footnotesize \{carsonzhao, jacobkong, fletchercai, hongfawang\}@tencent.com, wl2223@columbia.edu
}

\maketitle

\begin{abstract}
In this report, we propose a video-language pretraining~(VLP) based solution~\cite{kevin2022egovlp} for four Ego4D challenge tasks, including \nlq~(NLQ), \mq~(MQ), \OSCC~(OSCC), and \pnr~(PNR). Especially, we exploit the recently released Ego4D dataset~\cite{grauman2021ego4d} to pioneer \vlp~from pretraining dataset, pretraining objective, and development set. Based on the above three designs, we develop a pretrained video-language model that is able to transfer its egocentric video-text representation or video-only representation to several video downstream tasks. Our \vlp~achieves $10.46{R}@1\&IoU@0.3$ on NLQ, $10.33$ mAP on MQ, $74\%$ Acc on OSCC, $0.67$  sec error on PNR. The code 
is available at {\url{https://github.com/showlab/EgoVLP}}.
\end{abstract}

\section{Introduction}\label{sec:intro}
Video-Language Pretraining~(VLP) has prevailed in the regime of Vision~+~Language, aiming to learn strong and transferable video-language representation for powering a broad spectrum of video-text downstream tasks, video-text retrieval, video question answering, video-captioning.
The successes of VLP mainly stems from the availability of large-scale open-world video-text datasets such as \howto~\cite{miech2019howto100m}, which scrapes $134$K hours of instructional videos from the YouTube accompanied by text yielded from Automatic Speech Recognition.

Despite reaching an impressive data scale, videos in the existing video-text pretraining datasets~\cite{miech2019howto100m, bain2021frozen} are often of 3rd-person views and might have been edited before posting on the web.
Yet, there is a noticeable domain gap between the existing video-text pretraining datasets and 1st-person view videos such as those videos captured by wearable cameras or smart glasses.
Egocentric video has received increasing interests from academia~(e.g., activity anticipation~\cite{damen2022rescaling}) and industry (various applications in robotics and augmented reality).
But, due to such a domain gap, directly transferring the existing VLP models to egocentric downstream tasks cannot fully unleash the potential of large-scale pretraining approaches.
Roused by the favorable scale and diversity of recently released Ego4D~\cite{grauman2021ego4d} dataset, we are motivated to develop \vlp~models~\cite{kevin2022egovlp}, which can greatly benefit various egocentric video downstream applications.

In this report, we leverage our \vlp~\cite{kevin2022egovlp}~to a series of Ego4D challenge tasks, including one jointly video-text task: \nlq~(NLQ) and three video-only tasks: \mq~(MQ), \OSCC~(OSCC), and \pnr~(PNR).
We provide a general solution for VLP to tackle the above tasks and conduct a comprehensive analysis of the impact of different pretraining on tasks, e.g., without VLP, 3rd-person VLP, and 1st-person VLP.

\section{Approach}\label{sec:method}
\begin{figure*}[!t]
	\centering
	\includegraphics[width=1.0\linewidth]{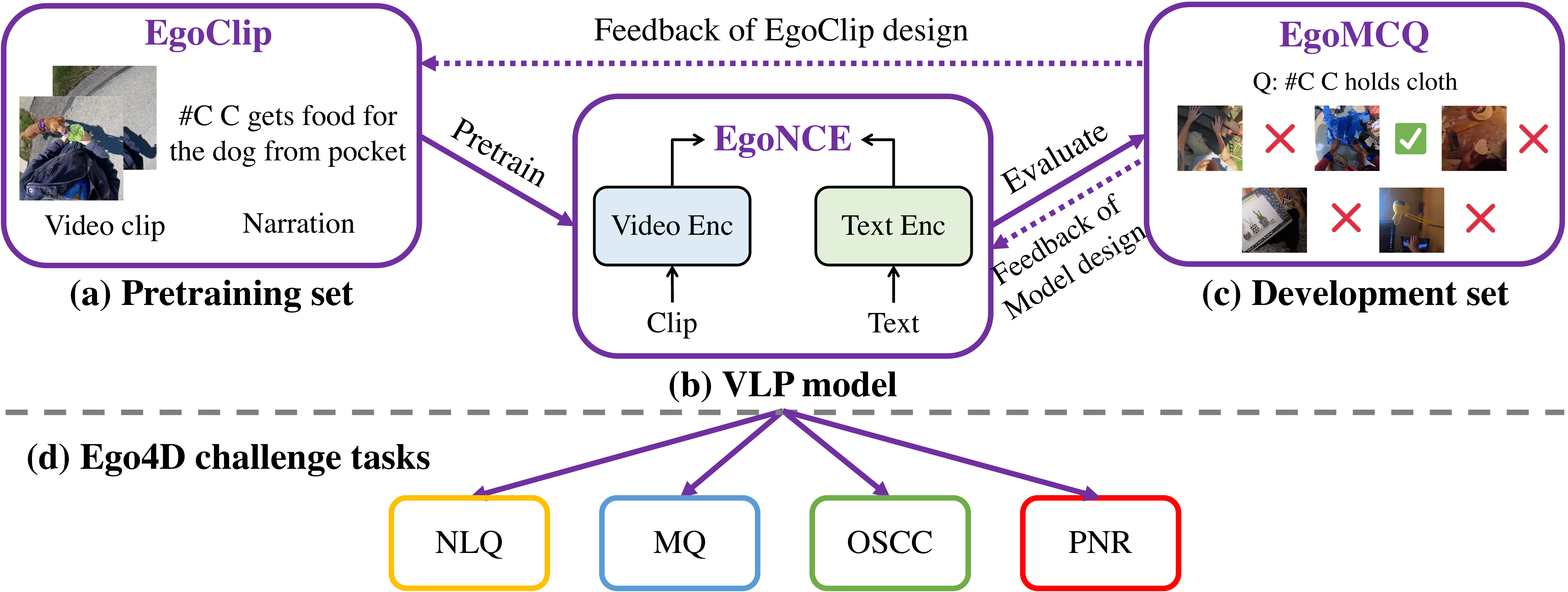}
	\vspace{-.5em}
	\caption{
	\textbf{Top:} Our \vlp~framework, which includes: (a) pretraining set \dataset; (b) VLP model; and (c) development set \eval. We use \dataset~to pretrain a VLP model with \model~ loss and then evaluate on \eval. According to the feedback, we iteratively refine our designs of (a) and (b). 
	\textbf{Down:}
	We transfer our pretrained model to four Ego4D challenge tasks, includes one jointly video-text task: NLQ, and three video-only tasks: MQ, OSCC, PNR.}
	\vspace{-1.5em}
	\label{framework}
\end{figure*}

\subsection{VLP Model}
We choose Frozen~\cite{bain2021frozen} as our pretraining architecture. As depicted in the Fig.~\ref{framework}(b), Frozen~\cite{bain2021frozen} design encompasses an elegant and simple dual encoder strategy (one per modality) which has favorable characteristics (e.g., indexability and efficiency~\cite{bain2021frozen}). 
Note that this allows the pretrained model for single-modality tasks (e.g., video-only tasks). 
In practice, the video encoder is a TimeSformer~\cite{timesformer} architecture while the text encoder builds upon DistillBERT~\cite{distilbert}. 
We adopt this notation: $(\mathcal{V}_i, \mathcal{T}_i)$ represent the video-test input to the model, while $\mathbf{v}_i$ and $\mathbf{t}_i$ are used to identify the L2 normalized video and text embedding with $d$ dimension.

\subsection{Egocentric Pretraining}
As illustrated in Fig.~\ref{framework}~\cite{kevin2022egovlp}, our pretraining framework includes three designs: \dataset, \model, and \eval.
We use \dataset~dataset for pretraining, which comprises $3.85$M video-text pairs well-chosen from Ego4D, covering a large variety of human daily activities.  Details please refer to~\cite{kevin2022egovlp}.
\textit{Notably, \dataset~has excluded videos that belong to the validation and test sets of Ego4D benchmark challenge.}
Next, we employ \model~as the model pretraining objective, which extends video-text InfoNCE~\cite{bain2021frozen} via positive and negative sampling strategies with formulation:
\begin{equation}
\mathcal{L}^\text{ego}=
\mathcal{L}^\text{ego}_\text{v2t}
+\mathcal{L}^\text{ego}_\text{t2v}.
\end{equation}

We formulate $\mathcal{L}^\text{ego}_\text{v2t}$ for simplicity whereas $\mathcal{L}^\text{ego}_\text{t2v}$ is defined in a symmetry way.
\begin{equation}
\begin{split}
	\mathcal{L}^\text{ego}_\text{v2t}=\frac{1}{| \mathcal{\widetilde{B}} |}\sum_{i\in\mathcal{\widetilde{B}}}  \log 
	\frac{
	{
	\sum_{k\in \mathcal{P}_i}\exp(\mathbf{v}_i^T\mathbf{t}_k /\tau)
	}
	}
	{  \sum_{j\in \mathcal{B}} \left( \exp(\mathbf{v}_i^T\mathbf{t}_j /\tau) +
	{\exp(\mathbf{v}_i^T\mathbf{t}_{j'} /\tau)}  \right) 
	},
	\label{egonce}
\end{split}
\end{equation}
where the numerator term corresponds to our proposed \textbf{action-aware positive samples}, which selects the positive sample within a batch by identifying narrations nouns and verbs. Then, batch samples that shared at least one noun and at least one verb are treated as positive samples:
$\mathcal{P}_i=\{j\in \mathcal{B}~|~\text{noun}(j)\cap\text{noun}(i)\neq\varnothing, \text{verb}(j)\cap\text{verb}(i)\neq\varnothing\}$.
While the denominator term corresponds to our proposed \textbf{scene-aware negative samples}.
For each video clip $i$, we sample an adjacent clip $i'\in \mathcal{N}(i)$, which is close to $i$ in time~(less than 1 min) within the same video.
Hence the batch is updated as $\mathcal{\widetilde{B}}=\{\underbrace{1,2,\cdots N}_{\mathcal{B}}, \underbrace{1',2',\cdots, N'}_{\mathcal{N}(\mathcal{B})} \}$. 
\model provides a general extension to adapt the existing VLP models for video-text pretraining datasets in the egocentric domain.

We evaluate our designs of \dataset~and \model on \eval, 
which contains 39K video-text multi-choices questions that are closer to pretraining domains and benchmark model video-text alignment, powering us to accurately validate and quickly iterate our decisions.

\subsection{Task-specific Transferring}
\begin{figure*}[!t]
	\centering
	\includegraphics[width=1.0\linewidth]{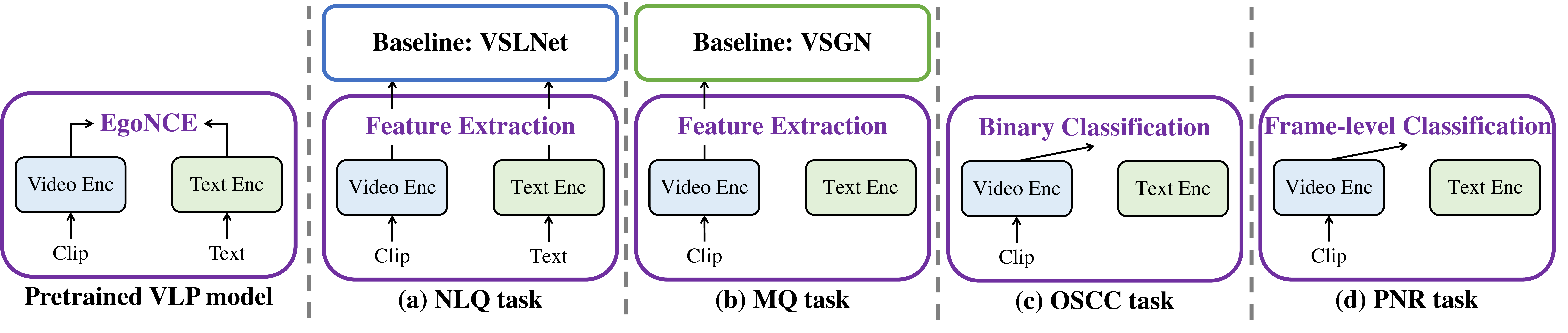}
	\vspace{-2em}
	\caption{Strategies for transferring pretrained VLP models to different Ego4D challenges.}
	\label{transfer}
\end{figure*}

In this section, we answer how to transfer pretrained VLP representations to multiple Ego4D challenge task, summarized in Fig.~\ref{transfer}.

\begin{figure*}[t]
 \begin{minipage}[t]{0.6\textwidth}
  \centering
  \scriptsize
  \setlength{\tabcolsep}{1.5mm}{
\begin{tabular}{clc|cccc}
	\toprule
	\textbf{Methods} & \multicolumn{2}{c|}{\textbf{Video-text Pre-extrated Features}} & \multicolumn{2}{c}{\textbf{IoU=0.3}} & \multicolumn{2}{c}{\textbf{IoU=0.5}} \\
	& Vis-text Enc & Vis-text PT & R@1 & R@5 & R@1 & R@5 \\  \midrule[1pt] 
	2D-TAN~\cite{zhang2020learning} &  SlowFast+BERT& - & $5.04$ & $12.89$ & $2.02$ & $5.88$   \\
	VSLNet~\cite{zhang2020span}  &  SlowFast+BERT   & - & $5.45$ & $10.74$ & $3.12$ & $6.63$  \\  
	 \midrule
 	VSLNet~\cite{zhang2020span}  &  Frozen       & \howto        & $3.95$ & $8.72$  & $2.01$ & $4.62$ \\	
	VSLNet~\cite{zhang2020span}  &  Frozen       & CC3M+WebVid2M & $5.06$ & $10.30$ & $2.71$ & $6.69$ \\		
	VSLNet~\cite{zhang2020span}  &  Frozen       & \dataset      & \underline{$10.53$} & \underline{$17.94$} & \underline{$5.96$} & \underline{$11.85$}\\
	VSLNet~\cite{zhang2020span}  &  Frozen+\model& \dataset      & $\mathbf{10.84}$ & $\mathbf{18.84}$ & $\mathbf{6.81}$ & $\mathbf{13.45}$ \\	
	\bottomrule
\end{tabular}
}\makeatletter\def\@captype{table}\makeatother\caption{Recall for several IoU on the NLQ task's val. set.}
  \end{minipage}
  \begin{minipage}[t]{0.3\textwidth}
   \centering
   \scriptsize
  \setlength{\tabcolsep}{1mm}{
\begin{tabular}{lc|cc}
	\toprule[1pt] 
	\textbf{Methods}  &  \textbf{Vis-Text PT} & \multicolumn{1}{c}{OSCC}  & \multicolumn{1}{c}{PNR}\\ 
	 & & Acc~(\%) & Err~(s)\\
	\midrule[1pt] 
    Always same                          &  - & $48.1$ & $1.032$ \\
	Bi-d LSTM &  ImageNet & $65.3$ & $0.790$  \\
	I3D ResNet-50     &  -        & $68.7$ & $0.739$ \\ 
	\midrule
	Frozen &   -            & $70.3$   & ${0.616}$           \\
	Frozen &  HowTo100M     & $71.7$   & $\mathbf{0.611}$           \\
	Frozen &  CC3M+WebVid2M & $71.5$   & $\underline{0.614}$           \\
	Frozen &  \dataset      & \underline{$73.4$} & ${0.618}$ \\
	Frozen+\model~ & \dataset  & $\mathbf{73.9}$ & $0.622$ \\
	\bottomrule[1pt] 
\end{tabular}
}
	\label{hands}
	 \makeatletter\def\@captype{table}\makeatother\caption{Performance on the OSCC~task and PNR~task's val set.}
\end{minipage}
\vspace{-1em}
\end{figure*}

\begin{table*}[thb]
\footnotesize
\centering
\setlength{\tabcolsep}{3pt}
\resizebox{\linewidth}{!}{
\begin{tabular}{clc|cccccc|cccccc}
		\toprule[1pt] 
		\textbf{Methods} & \multicolumn{2}{c|}{\textbf{Video Pre-extracted Features}} & \multicolumn{2}{c}{\textbf{IoU=0.3}} & \multicolumn{2}{c}{\textbf{IoU=0.5}} &
		\multicolumn{2}{c|}{\textbf{IoU=0.7}} &
		\multicolumn{4}{c}{\textbf{mAP~(\%) @ IoU}}\\
		& Vis Enc & Vis-text PT & R@1 & R@5 &R@1 & R@5 & R@1 & R@5 & 0.1 & 0.3 & 0.5 & Avg \\ 
		\midrule[1pt] 
		VSGN~\cite{zhao2021video} & SlowFast    &  -        & $33.45$ & $58.43$ & $25.16$ & $46.18$ & $15.36$ & $25.81$ & $9.10$ & $5.76$ & $3.41$ & $6.03$  \\\midrule[1pt] 
		VSGN~\cite{zhao2021video} & Frozen & \howto         & $31.40$ & $52.61$ & $22.28$ & $41.29$ & $13.41$ & $23.21$ & $9.83$ & $6.72$ & $3.84$ & $6.72$  \\		
		VSGN~\cite{zhao2021video} & Frozen  &CC3M+WebVid2M  & $32.08$ & $56.40$ & $23.46$ & $43.81$ & $13.73$ & $23.77$ & $9.83$ & $6.40$ & $3.86$ & $6.58$  \\		
		VSGN~\cite{zhao2021video} & Frozen & \dataset& \underline{$40.06$} & \underline{$63.71$} & \underline{$29.59$} & \underline{$48.32$} & \underline{$17.41$} & \underline{$26.33$} & \underline{$15.90$} & \underline{$10.54$} & \underline{$6.19$} & \underline{$10.69$} \\	
		VSGN~\cite{zhao2021video} & Frozen+\model~ & \dataset& $\mathbf{40.43}$ & $\mathbf{65.67}$ & $\mathbf{30.14}$ & $\mathbf{51.98}$ & $\mathbf{19.06}$ & $\mathbf{29.77}$ &  $\mathbf{16.63}$ & $\mathbf{11.45}$ & $\mathbf{6.57}$ &$\mathbf{11.39}$ \\
		\bottomrule[1pt] 
	\end{tabular}
}
\caption{Recall and mAP metrics for several IoU on the Moment Query task's val. set.}
\label{mq}
\vspace{-1em}
\end{table*}

\textbf{\nlq.}
This task is a kind of video-text localization, a jointly video-text task. 
The clip in this dataset tends to be long~($480$~sec on average), so it is difficult to achieve end-to-end fine-tuning. 
Thus, we propose to evaluate offline video-text features. 
Especially, we adopt the official baselines VSLNet~\cite{zhang2020span}, which takes $2304$ dim SlowFast features~($1.87$ fps, Kinetics 400 pretrained) and $768$ dim BERT features as input,
and we substitute them with output features of pretrained VLP video and text encoders to validate the pretraining effectiveness, as depicted in Fig.~\ref{transfer}~(a).

\textbf{\mq.}
This task is a kind of temporal action localization, a video-only task. 
Similar to the NLQ task, we replace the input Slowfast features of baseline VSGN~\cite{zhao2021video} with extracted video features of our VLP for evaluation, as depicted in Fig.~\ref{transfer}~(b).

\textbf{\OSCC.}
This task is formulated as a two-way action classification, a video-only task.
As shown in Fig.~\ref{transfer}~(c), We set the output embedding dimension of pretrained video encoder as $2$ and fine-tune it with cross-entropy loss.

\textbf{\pnr.}
This task can be regarded as the frame-level action classification, a video-only task. 
As shown in Fig.~\ref{transfer}~(d), similar to OSCC, we change the embedding dimension of our pretrained video encoder to $16$ and fine-tune it with cross-entropy loss.

\section{Experiments}\label{sec:exps}
\subsection{Implementation Details}
Following the settings of official Frozen~\cite{bain2021frozen}\footnote{https://github.com/m-bain/frozen-in-time}, the video encoder is initialized with ViT~\cite{dosovitskiy2020image} weights trained on ImageNet-21K with sequence dimension $768$. 
The text encoder is based on huggingface's $\texttt{distilbert-base-uncased}$.
During pretraining, the dimension of common feature space is set as $256$, and the temperature parameter $\tau$ is set to $0.05$. Each video is resized to $224\times 224$ as input with sample frames number $4$ and batch size $512$.
We use the Adam optimizer with a learning rate of $3\times 10^{-5}$ with a total epoch of $10$. 
When transferring to downstream tasks, we select the checkpoints with the best score on \eval~benchmark.
For NLQ and MQ tasks, we extract the video features with fps $1.87$ and sampling frame number $4$ with stride $4$.
In the fine-tuning stage, we keep the default setting of baselines\cite{zhang2020span, zhao2021video}.
For OSCC and PNR tasks, we sample each clip with $16$ frames as input and set the epoch equal to $10$. 
And we adopt the same settings of pretraining e.g. learning rate. 

\begin{figure*}[t]
 \begin{minipage}[t]{0.55\textwidth}
  \centering
  \scriptsize
  \setlength{\tabcolsep}{1.5mm}{
\begin{tabular}{clc|cccc}
	\toprule
	\textbf{Methods} & \multicolumn{2}{c|}{\textbf{Video-text Pre-extrated Features}} & \multicolumn{2}{c}{\textbf{IoU=0.3}} & \multicolumn{2}{c}{\textbf{IoU=0.5}} \\
	& Vis-text Enc & Vis-text PT & R@1 & R@5 & R@1 & R@5 \\  \midrule[1pt] 
	VSLNet  &  SlowFast+BERT   & - & $5.47$ & $11.21$ & $2.80$ & $6.57$  \\  
	 \midrule
 	VSLNet  &  Frozen       & \howto        & $3.77$ & $6.87$  & $1.62$ & $3.45$ \\	
	VSLNet  &  Frozen       & CC3M+WebVid2M & $4.87$ & $8.67$ & $2.50$ & $4.97$ \\		
	VSLNet  &  Frozen       & \dataset      & \underline{$10.34$} & \underline{$15.81$} & \underline{$6.24$} & \underline{$10.39$}\\
	VSLNet  &  Frozen+\model& \dataset      & $\mathbf{10.46}$ & $\mathbf{16.76}$ & $\mathbf{6.24}$ & $\mathbf{11.29}$ \\	
	\bottomrule
\end{tabular}
}\makeatletter\def\@captype{table}\makeatother\caption{Performance on the NLQ task's test set.}
  \end{minipage}
  \begin{minipage}[t]{0.35\textwidth}
   \centering
   \scriptsize
  \setlength{\tabcolsep}{1mm}{
{
\begin{tabular}{clc|cc}
	\toprule
	\textbf{Methods} & \multicolumn{2}{c|}{\textbf{Video-text Pre-extrated Features}} & \multicolumn{1}{c}{\textbf{IoU=0.5}} & \multicolumn{1}{c}{\textbf{mAP$(\%)$IoU}} \\
	& Vis Enc & Vis-text PT & R@1  & Avg \\  \midrule[1pt] 
	VSGN &  SlowFast & - & $24.25$  & $5.68$   \\ \midrule
	VSGN & Frozen & \howto&  $18.06$  & $5.28$   \\
	VSGN & Frozen & CC3M+WebVid2M  & $19.74$  & $5.95$   \\
	VSGN & Frozen & \dataset  & \underline{$27.98$}  & \underline{$9.78$}   \\
	VSGN & Frozen+\model & \dataset & $\mathbf{28.03}$  & $\mathbf{10.33}$   \\
	\bottomrule
\end{tabular}
}
}
	\label{hands}
	 \makeatletter\def\@captype{table}\makeatother\caption{Performance on the MQ task's test set.}
\end{minipage}
\vspace{-1em}
\end{figure*}

\subsection{Results}
\textbf{\nlq.}
We report the NLQ validation results on Tab.~1. 
We observe a large boost in performance offered by our pretrained model for all metrics. 
Notably, we improve R@1 for IoU=0.3 from $5.45$ to $10.84$, despite our video branch not being pre-trained on Kinetics400.
Besides, we significantly surpass VLP pretrained on \ccweb~and~\howto.
We believe this increase is due to the egocentric data availability and the video-text interaction learned from large-scale egocentric pretraining. 
In Tab~4, we display the test set.

\textbf{\OSCC.}
We report the OSCC validation results on Tab.~2. Once again, our model achieves the best performance of all baselines, $2.4\%$ than \ccweb~counterparts, which indicates our visual representations are able to focus on the fine-grained clues related to changes. We select the best Frozen+EgoNCE variant and evaluate on the test set, and get $73.7\%$ accuracy.

\textbf{\pnr.}
We report the PNR validation results on Tab.~2 and found that the pretraining effect is minor on this task. 
We select the Frozen+EgoNCE variant and evaluate on the test set, and get $0.666$~s localization error.

\textbf{\mq.}
We report the MQ validation results in Tab.~3,
We find that our features achieves the best performance over SlowFast features with an increase of $4.66\%$ in Avg mAP.
Moreover, we maintain better performance with respect to 3rd-person large-scale pretraining datasets.
This demonstrates that the 1st-person VLP model also learns competitive video representations. 
In Tab~5, we display the MQ test set result of different VLP variants.

\section{Conclusion and Limitations}
We present an egocentric video-language pretraining solution~\cite{kevin2022egovlp} for four Ego4D challenge tasks, including NLQ, MQ, OSCC, and PNR.
Specifically, we provide a general solution for VLP to tackle the above challenges and conduct extensive experiments to analyze the impact of different pretraining on different tasks. And we demonstrate the superiority of our \vlp~on three tasks

\textbf{Limitations:}
VLP requires a large training cost ($1,536$ GPU hrs for our model) and may be limited by the model architecture thus not flexible for a specific task.

{\small
\bibliographystyle{ieee_fullname}
\bibliography{Ego4D}
}

\end{document}